%% file: main.tex
\documentclass[conference]{IEEEtran}
\IEEEoverridecommandlockouts
\usepackage{cite}
\usepackage{amsmath,amssymb,amsfonts}
\usepackage{algorithmic}
\usepackage{graphicx}
\usepackage{textcomp}
\usepackage{comment}
\usepackage[dvipsnames]{xcolor}
\usepackage{subcaption}
\usepackage{multirow}
\usepackage[export]{adjustbox}
\def\BibTeX{{\rm B\kern-.05em{\sc i\kern-.025em b}\kern-.08em
    T\kern-.1667em\lower.7ex\hbox{E}\kern-.125emX}}
\begin{document}

\title{Benchmarking Edge AI Platforms for High-Performance ML Inference\\
}

 \author{\IEEEauthorblockN{Rakshith Jayanth$^*$, Neelesh Gupta$^*$, Viktor Prasanna}
 \IEEEauthorblockA{\textit{Ming Hsieh Department of Electrical and Computer Engineering} \\
 \textit{University of Southern California}\\
 Los Angeles, USA \\
 \{jayanthr, neeleshg, prasanna\}@usc.edu}}

\maketitle
\begingroup\renewcommand\thefootnote{* }
\footnotetext{These authors contributed equally.}
\endgroup


\begin{abstract}
Edge computing's growing prominence, due to its ability to reduce communication latency and enable real-time processing, is promoting the rise of high-performance, heterogeneous System-on-Chip solutions. While current approaches often involve scaling down modern hardware, the performance characteristics of neural network workloads on these platforms can vary significantly, especially when it comes to parallel processing, which is a critical consideration for edge deployments. To address this, we conduct a comprehensive study comparing the latency and throughput of various linear algebra and neural network inference tasks across CPU-only, CPU/GPU, and CPU/NPU integrated solutions.
{We find that the Neural Processing Unit (NPU) excels in matrix-vector multiplication (58.6\% faster) and some neural network tasks (3.2$\times$ faster for video classification and large language models). GPU outperforms in matrix multiplication (22.6\% faster) and LSTM networks (2.7$\times$ faster) while CPU excels at less parallel operations like dot product. NPU-based inference offers a balance of latency and throughput at lower power consumption. GPU-based inference, though more energy-intensive, performs best with large dimensions and batch sizes.
We highlight the potential of heterogeneous computing solutions for edge AI, where {diverse} compute units can be strategically leveraged to boost accurate and real-time inference.
}

\end{abstract}


\section{Introduction}

\input{content/1_introduction}

\section{Background}
\input{content/2_background}

\section{Related Works}
\input{content/3_related_works}

\section{Evaluation}

\input{content/4_benchmarking_methodology}

\section{Results}

\input{content/5_experimentation}

\section{Conclusion}

\input{content/6_conclusion}

\section*{Acknowledgment}
This work was supported by Army Research Office (ARO) under award number W911NF2220159 and National Science Foundation (NSF) under grants under award numbers SaTC-2104264 and CNS-2009057.

\bibliographystyle{IEEEtran}
\bibliography{references}

\end{document}

%% file: content/1_introduction.tex
The rapid expansion of Artificial Intelligence (AI) applications has intensified the need for inference capabilities at the edge~\cite{xu2021edge, li2024flexnn, liu2022survey, xu2020edge, shi2016edge, zhou2019edge, wang2020convergence}, or in resource-constrained environments. Edge computing brings data processing closer to the source, reduces latency, enhances privacy, and enables real-time decision-making. This shift towards edge AI has become crucial in various domains, including autonomous vehicles~\cite{garg2018uav}, smart cities~\cite{khan2020edge}, industrial IoT~\cite{kong2022edge, mahdavinejad2018machine}, and mobile devices~\cite{baller2021deepedgebench}, where rapid and efficient on-device inference is essential.

\begin{figure}
    \centering
    \includegraphics[width=0.5\textwidth]{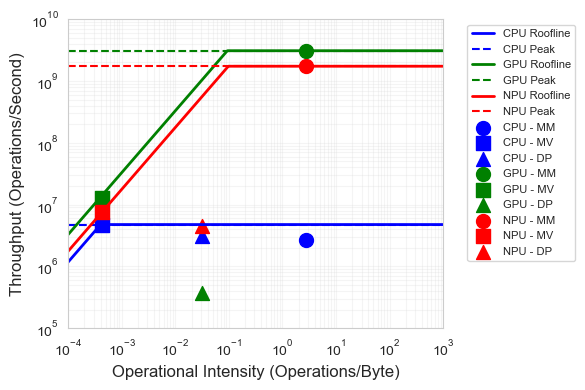}
    \caption{Roofline model comparing performance of CPU, CPU/GPU, and CPU/NPU architectures for linear algebra operations. The plot shows a consistent performance hierarchy: CPU (lowest), CPU/NPU (middle), and CPU/GPU (highest) across both memory-bound and compute-bound regions.}
    \label{fig:enter-label}
\end{figure}

To meet these demands, semiconductor manufacturers are swiftly developing increasingly complex System-on-Chip (SoC) platforms that integrate traditional CPUs with specialized accelerators such as GPUs and NPUs. Exemplary designs include Intel's Core Ultra Processor~\cite{gomes2022meteor}, also known as Intel AIPC, with an integrated NPU, Apple's M-series chips~\cite{banerjee2018microarchitectural} with dedicated neural engines, and Qualcomm's Snapdragon platforms~\cite{martinez2022smartphone} featuring AI optimized Digital Signal Processors. 
These advancements significantly boost on-device AI performance and energy efficiency, expanding computing capabilities at the edge.

As sophisticated SoCs become more prevalent, comprehensive benchmarking becomes crucial for system designers, software developers, and end-users to maximize hardware capabilities. 
Benchmarking exposes processing unit strengths, guiding hardware selection and workload optimization on heterogeneous platforms.
This study focuses on benchmarking two critical aspects of Machine Learning (ML) workloads: fundamental linear algebra operations and Deep Neural Network (DNN) applications. Linear algebra operations, such as matrix multiplication,~\cite{mummidi2024highly, hu2024enhancing} form the backbone of many ML algorithms and aid in understanding the raw computational capabilities of different compute platforms. DNN applications, including video classification~\cite{rehman2023deep} and text generation~\cite{zhang2024tinyllamaopensourcesmalllanguage}, represent real-world AI tasks that leverage these fundamental operations in complex ways. By evaluating performance across these workloads, we provide a comprehensive view of how different compute units handle the diverse computational requirements of modern AI applications at the edge.

Our main contributions are summarized as follows:

\begin{itemize}
    \item We conduct a benchmark study that analyses the performance of fundamental linear algebraic operations and neural network models on an integrated heterogeneous platform consisting of CPU, GPU and NPU.
    \item Our results show that NPU excels in matrix-vector multiplication, reducing latency by 58.54\% compared to GPUs. Conversely, GPU outperforms in matrix multiplication, offering 22.6\% lower latency and 2$\times$ higher throughput than NPU. For dot product, CPU demonstrates the lowest latency among all platforms.
    \item In model inference, NPU outperforms GPU by 3.2$\times$ for LLM tasks, while GPU surpasses NPU by 2.7$\times$ for LSTM models. For video classification, NPU performance remains consistent across batch sizes, whereas GPU performance scales, favoring larger batches.
    \item We conclude that NPU excels in LLM inference and memory-bound operations such as matrix-vector multiplication due to efficient DMA (Direct Memory Access) utilization, while GPU is more efficient for LSTM models and compute-bound operations such as matrix multiplication. CPU performs best for low-complex compute operations such as dot-product as they avoid additional system memory accesses. NPU offers consistent performance across varied batch sizes, but GPU is superior for large-scale batch processing.
\end{itemize}

%% file: content/2_background.tex
Our effort focuses on benchmarking applications using integrated heterogeneous platforms. An integrated heterogeneous platform, in simple terms, is a combination of accelerator platforms implemented as a System-on-Chip (SoC). For this work, we consider three key components: CPU, GPU, and NPU (Neural Processing Unit). In the following section, we provide an overview of NPU and explain the interactions between these {devices}. Understanding these interactions is crucial as it forms the foundation for our benchmarking efforts, allowing us to comprehensively evaluate performance across the diverse computational resources available in modern SoCs.

\begin{figure}[!th]
    \centering
    \includegraphics[width=0.95\linewidth]{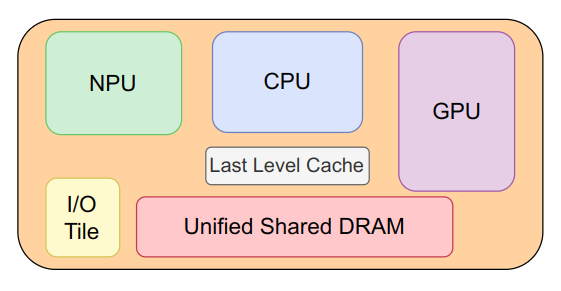}
    \caption{Integrated Heterogeneous System}
    \label{Interated_system}
\end{figure}

\subsection{Neural Processing Unit}
NPUs~\cite{jang2021saprsity} are specialized processors designed specifically for AI workloads, excelling particularly at neural network inference and training. Their integration into modern SoCs has become increasingly common. NPUs offer superior energy efficiency compared to both CPUs and GPUs, making them primary targets for edge computing environments where power constraints are critical. A key feature of NPU architectures is their ability to exploit quantization, with many modern designs supporting mixed-precision arithmetic to balance performance and accuracy. Typically based on systolic array architectures, NPUs provide massive parallelism and maximize data reuse, enabling highly efficient execution of AI algorithms. These characteristics position NPUs as crucial components in the evolving landscape of heterogeneous computing for AI applications.
Since our operations are executed on Intel NPU, here is a brief architectural overview of the same. Intel NPU is integrated into the Intel AIPC, and the compute acceleration is supported by scalable NCEs (Neural Compute Engines). These NCEs have AI acceleration blocks that perform matrix operations efficiently. The NPU is equipped with SHAVE (Streaming Hybrid Architecture Vector Engines), which performs parallel computing on general-purpose tasks. DMA (Direct Memory Access) engines, which have been integrated into NPU, perform efficient transfer of data from system memory to software-managed cache.

\begin{figure}[!th]
    \centering
    \includegraphics[width=0.95\linewidth]{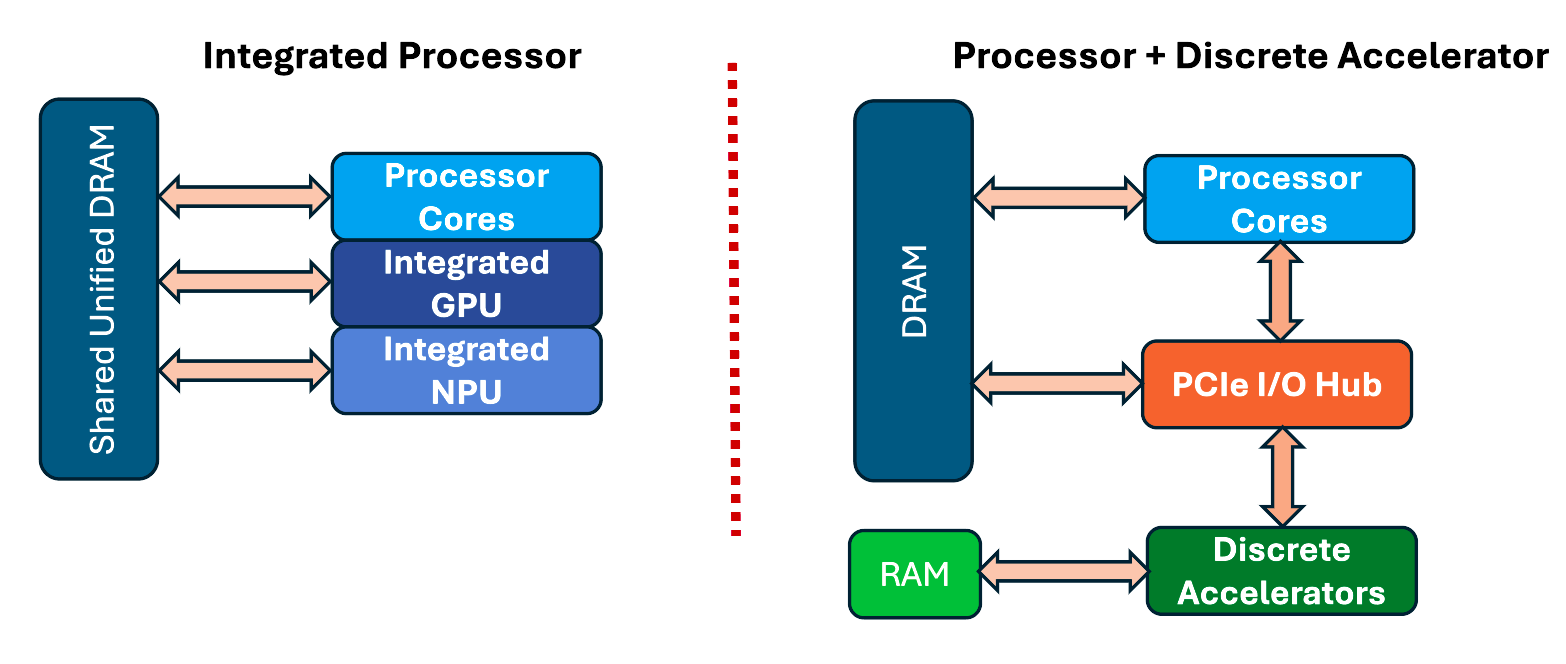}
    \caption{System Interaction With Integrated Accelerators and Discrete Accelerators}
    \label{system_interaction_fig}
\end{figure}

\subsection{Interaction Between The Compute Platforms}

Understanding the interaction between CPU, GPU, and NPU is crucial as it forms the basis for our benchmarking results. Most modern SoCs, such as Intel AIPC, have adopted a shared memory architecture unified across these three {devices} as shown in Fig. \ref{system_interaction_fig}. In this unified memory system, each platform communicates by writing to corresponding shared spaces in memory. This approach offers a major advantage: seamless communication between platforms without unnecessary data transfers. In contrast, systems with discrete accelerators like NVIDIA GPUs require data transfer from the SoC's memory to the accelerator's memory, incurring significant latency. However, integrated processors have a disadvantage in terms of limited memory bandwidth, primarily due to CPU constraints. Discrete accelerators, on the other hand, can leverage HBMs (High Bandwidth Memories), offering potentially higher throughput for memory-intensive tasks. These architectural differences significantly impact performance and efficiency in heterogeneous computing environments.

%% file: content/3_related_works.tex
Recent benchmarking studies of linear algebra operations and neural network inference on heterogeneous platforms have advanced significantly. 
This progress is driven by increasingly complex edge computing devices and growing demand for efficient edge AI processing.
Most studies have focused on evaluating performance across CPUs, GPUs, and specialized accelerators such as NPUs. These investigations discuss the strengths and limitations of the hardware. They analyze performance, energy efficiency, memory usage, and scalability across various workloads. The following sections review key contributions in the benchmarking of fundamental linear algebra operations and neural network models.

\subsection{Linear Algebra}
{
There exist multiple works on analyzing linear algebra operations, specifically BLAS (Basic Linear Algebra Subprograms) operations, on heterogeneous platforms. Works on analyzing BLAS operations~\cite{anderson2012predictive, abdelfattah2020matrix} concentrate on performance analysis on GPU architectures and focus on evaluation of large matrices. The analysis is performed as part of model development for performance improvement and is conducted on GPUs. Other dedicated works on benchmarking BLAS operations~\cite{nath2010improved, Gautier2021Evaluation, Wang2016BLASX} concentrate their efforts on level-3 BLAS operations such as matrix multiplication and use high-performing GPUs as their platform. 

However, all the aforementioned works make use of discrete and high-performing GPUs and not integrated GPUs, nor do they evaluate any operations on NPUs. Additionally, these works do not perform a full-scale analysis of all levels of BLAS operations as they only work with level-3 operations. Therefore, their analysis is not applicable to understanding the performance of these operations on edge devices that comprise CPUs, integrated GPUs, and NPUs.
In contrast, our work performs a full-scale analysis of all levels of BLAS operations on CPU, GPU, and NPU {devices} that are integrated into a single-chip edge system.
}

\subsection{Neural Networks}
{
Comprehensive analysis of neural network models has seen a lot of engagement in prior works on various platforms, including edge devices. Notably, detailed benchmark analysis of DNN models~\cite{almeida2019embench, bianco2018benchmark, hadidi2018distributed} was performed on high-end CPU and GPU architectures. However, this analysis is not suited for edge applications as edge {systems} typically do not deal with high-performing GPUs and CPUs. We also find prior works that work on the analysis of ML workloads performed on Android devices~\cite{luo2019aiot} that succeed in making an overall performance comparison among models but do not discuss platform heterogeneity. Other prior works on android~\cite{hanhirova2018latency}, though considering the heterogeneous performance on CPU and GPU, concentrate on specific workloads such as CNN and do not evaluate performance on NPU. Analysis of workloads on smartphones~\cite{ignatov2018ai} not only executed on CPU and GPU workloads but also included performance evaluation on NPU. However, the NPU used in the analysis did not support quantization. Though the more recent work on benchmarking effort~\cite{baller2021deepedgebench} accounts for performance on all the platforms, it does not establish the relation between the performance of the models and the fundamental linear algebraic operations.

On the other hand, our work not only gives a comprehensive analysis of neural network models across integrated compute units but also tries to establish its relation with the performance of fundamental linear algebraic operations. To the best of our knowledge, our work is one of the first to present a complete analysis of benchmarking fundamental linear algebraic operations and ML models on edge platforms. 
}

%% file: content/4_benchmarking_methodology.tex
\begin{table*}[h]
\centering
\caption{Suite of Linear Algebra Tasks}
\begin{tabular}{|c|c|c|}
\hline
\textbf{Task} & \textbf{BLAS Level} & \textbf{Description} \\
\hline
Matrix Multiplication & 3 & Multiply two matrices $\textbf{A}$ ($N \times N$) and $\textbf{B}$ ($N\times N$). \\
\hline
Matrix-Vector Multiplication & 2& Multiply a matrix $\textbf{A}$ ($N \times N$) with a vector $\textbf{x}$ ($N \times 1$). \\
\hline
Dot Product & 1 &Compute the dot product of two vectors $\textbf{x}$ and $\textbf{y}$, both of length $N$.  \\
\hline
\end{tabular}
\label{tab:LA_ops}
\end{table*}

\begin{table*}[h]
\centering
\caption{Suite of Neural Network Inference Tasks}
\begin{tabular}{|c|c|c|c|c|}
\hline
 \textbf{Model} & \textbf{Input Type}  & \textbf{Parameters} & \textbf{Description} \\
\hline
MobileNetV2 ~\cite{sandler2018mobilenetv2} & Video & 6.9M &Efficient CNN for mobile and embedded vision applications.\\
\hline
LSTM ~\cite{hochreiter1997long} & Time series & 1.6M & Processes input in the forward direction. Captures past context.\\
\hline
TinyLlama ~\cite{zhang2024tinyllamaopensourcesmalllanguage}& Text & 1.1B & Compact transformer-based language model using grouped-query attention. \\

\hline
\end{tabular}
\label{tab:NN_ops}
\end{table*}

We benchmark key linear algebra operations and neural network models for inference. We are using Intel AIPC, which provides an integrated heterogeneous platform consisting of CPU, GPU, and NPU. Our benchmarking operation involves the use of multiple software frameworks that yield optimized results on Intel devices. In this section, we will provide a brief overview of these software frameworks and their workflows. Additionally, we will discuss the tasks involved along with the performance metrics we use for evaluation.

\begin{table}[htb!]
  \caption{Processor Specifications}
  \resizebox{\linewidth}{!}{\begin{tabular}{|c|c|c|c|}
    \hline
    \textbf{Feature} & \textbf{CPU} & \textbf{GPU} & \textbf{NPU}  \\
    \hline
    Operating Frequency & 5 GHz & 2.3 GHz & 1 GHz\\
    Power & 35 - 65 W & 75 W & 35 W\\
    Execution Units (EUs) & 16 Cores & 128 EUs & 4096 MAC Units\\
    \hline 
  \end{tabular}}
  \label{tab:processor_spec}
\end{table}


\begin{figure}[!th]
    \centering
    \includegraphics[width=0.95\linewidth]{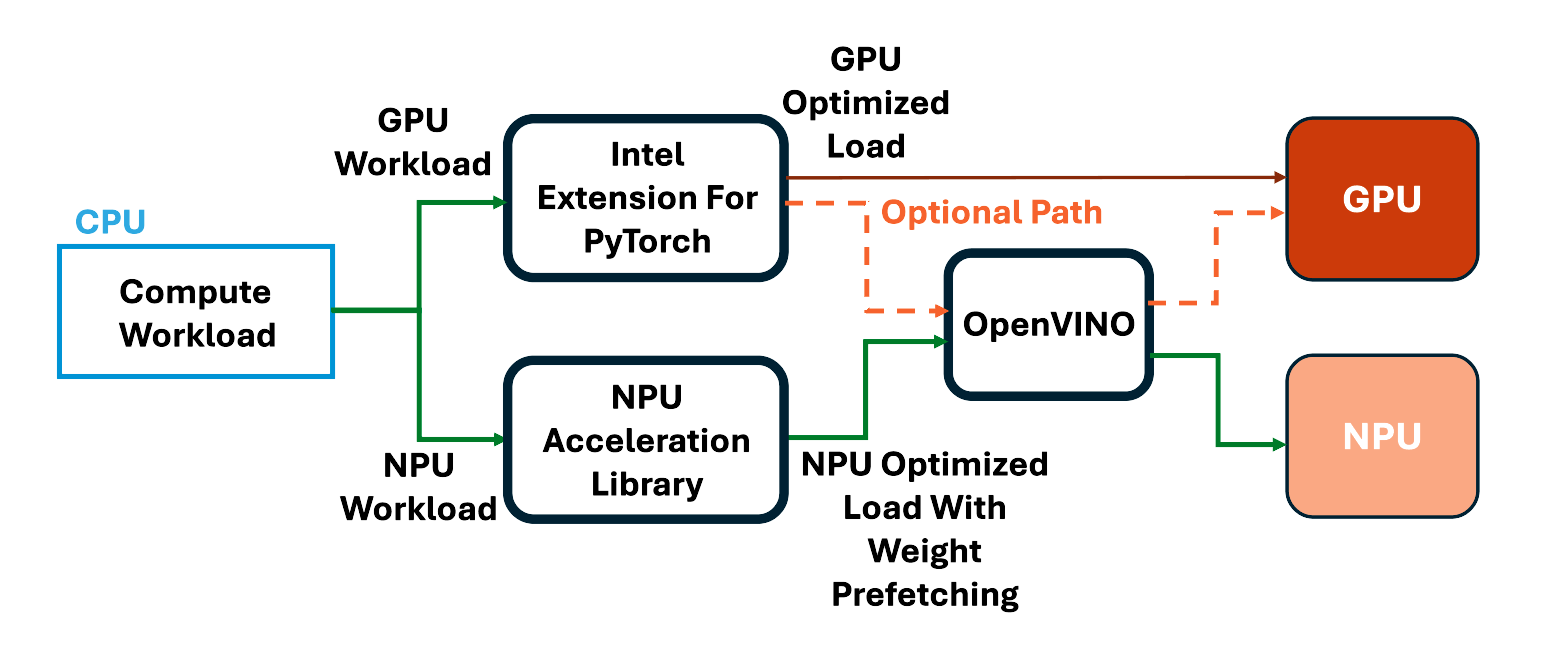}
    \caption{Workflow diagram of Intel frameworks illustrating coordination and workload distribution.}
    \label{workflow_fig}
\end{figure}

\subsection{Software Framework}

For our experiments, we used two frameworks offered by Intel: OpenVINO and IPEX (Intel Extension for Pytorch). Additionally, for computations on NPU, we utilized Intel's NPU acceleration library, which is based on OpenVINO. The workflow of these frameworks is illustrated in Fig. \ref{workflow_fig}. The OpenVINO framework coordinates communication and workload distribution across all memory and compute platforms. IPEX, an independent framework, can be used independently or in conjunction with OpenVINO. It offers optimizers that apply Intel device-specific optimizations to PyTorch models and distributes workloads between CPU and GPU by performing dynamic task scheduling. Both IPEX and OpenVINO implement lazy evaluation, which ensures that operations are executed only when results are required, thus reducing unnecessary memory access. Also, they ensure that operations on different platforms are executed asynchronously and tasks are scheduled dynamically. The NPU acceleration library optimizes NPU usage by exploiting its features of weight prefetching and efficient DMA accesses between cache and system memory, along with maximizing compute utilization.

\begin{table*}
  \caption{Performance of Linear Algebra Primitives (N=1024)}
  \resizebox{\textwidth}{!}{\begin{tabular}{|c||c|c|c||c|c|c||c|c|c|}
    \hline
    \multicolumn{1}{|c||}{} & \multicolumn{3}{|c||}{\textbf{Matrix Multiplication}} & \multicolumn{3}{|c||}{\textbf{Matrix-Vector Multiplication}} & \multicolumn{3}{|c|}{\textbf{Dot Product}} \\
    \cline{2-10}
    \textbf{Metric} & \textbf{CPU} & \textbf{GPU} & \textbf{NPU} & \textbf{CPU} & \textbf{GPU} & \textbf{NPU} & \textbf{CPU} & \textbf{GPU} & \textbf{NPU}\\
    \hline
    Latency (ms) & 1219.00  & \textbf{0.80} &1.04  & 0.86 & 0.20 &  \textbf{0.08} & \textbf{0.17} & 19.40 & 0.52\\
    Avg Throughput (GFLOPS) & 2.56  & \textbf{2947.37} & 1653.65 & 4.54 & \textbf{12.42} & 7.14 &  2.94 & 0.36 & \textbf{4.32}\\
    \hline 
  \end{tabular}}
  \label{tab: LA_performance}
\end{table*}

\subsection{Benchmarking Linear Algebra}
We analyze three fundamental linear algebraic operations: dense matrix multiplication, dense matrix-vector multiplication, and vector-vector dot product (Table \ref{tab:LA_ops}). {The motivation for the choice of these operations was based on the fact that they are the core operations in the neural network models. Analysis of these results helps extrapolate the results obtained for neural network models. Along with this, these operations form a comprehensive set evaluating all levels of BLAS operation.} 
For matrix and matrix-vector multiplication, we use square matrices of size $N\times N$, with $N$ ranging from 16 to 1024, typically focusing on 512 to 1024.
We exclusively use 16-bit floating point (FP-16) operations, as NPU only supports this precision, ensuring uniform analysis.

\noindent\textbf{Performance Metrics.} The metrics we use for these operations are latency and average throughput.

\subsection{Benchmarking Neural Network Inference}
We analyze the performance of a CNN-based model for video classification, an LSTM model, and a decoder-only transformer model for LLM inference, as mentioned in Table \ref{tab:NN_ops}. {The computational diversity, along with differences in the regularity of memory accesses, makes these models ideal candidates for benchmarking across various hardware platforms. Specifically, the classification model (MobileNetV2) demonstrates the impact of batch processing, the LSTM model provides insights into performance with irregular memory accesses, and the LLM model (LLaMA) demonstrates the interplay between different operations, as its prefill stage is dominated by matrix multiplications while the repeated decode stages primarily consist of matrix-vector operations.} For video classification analysis, we analyzed performance by varying batch sizes. Along with floating point operations, we measure the performance on quantized INT-8 weights as well. Execution on NPU took place with the OpenVINO framework for video classification, while we used the NPU library constructs for the LSTM and LLM models. Similar to linear algebraic operations, FP-16 was used across all platforms for analyzing the neural network models. 

\noindent\textbf{Performance Metrics.} For LSTM and LLM models, \textbf{inference latency} will be the \textbf{performance metric}, and for video classification, along with inference latency, \textbf{FPS (Frames Per Second)} will be used as a measure of throughput. Inference latency refers to the total time a model takes to generate output from when an input is provided. FPS is the number of individual images that a device can process in one second.

%% file: content/5_experimentation.tex
\begin{table}
  \centering

  \caption{Inference Latency (ms) of MobileNetV2}
  \begin{tabular}{|c||c|c||c|c|}
    \hline
    \multicolumn{1}{|c||}{} & \multicolumn{2}{|c||}{\textbf{FP-16 Weights}} & \multicolumn{2}{|c|}{\textbf{INT-8 Weights}}\\
    \cline{2-5}
    \textbf{Batch Size} & \textbf{GPU} & \textbf{NPU} & \textbf{GPU} & \textbf{NPU}\\
    \hline
    1 & 4.95 & \textbf{1.73} & 5.37 & \textbf{1.33} \\
    2 & 4.51 & \textbf{1.61}& 2.70 & \textbf{1.37} \\
    4 & 2.21 & \textbf{1.49} & 1.48 & \textbf{1.26} \\
    8 & \textbf{1.52} & 1.59 & \textbf{0.90} & 1.18 \\
    16 & \textbf{1.20} & 1.35 & \textbf{0.75} & 1.16 \\
    32 & \textbf{1.12} & 1.49 & \textbf{0.61} & 1.26 \\
    \hline 
  \end{tabular}
  \label{tab: Mobilenet_performance}
\end{table}

\begin{figure}[!h]
    \centering
    \includegraphics[width=\linewidth]{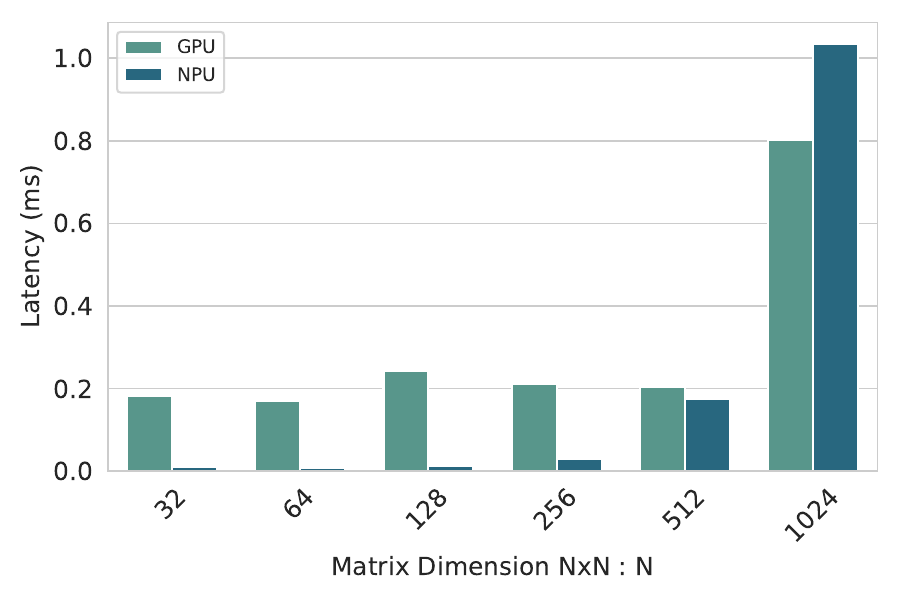}
    \caption{Latency for Matrix Multiplication}
    \label{latency_fig_la_mm}
\end{figure}

In this section, we evaluate the performance of three linear algebra primitives and three neural network models that were chosen based on their computational and model diversity, respectively, on the integrated CPU, GPU, and NPU {compute} platform of Intel AIPC.

 With respect to matrix multiplication, as shown in Table \ref{tab: LA_performance} and Fig. \ref{latency_fig_la_mm}, the NPU demonstrates superior performance for smaller matrices compared to the GPU. However, as matrix sizes increase, GPU computation latency is significantly reduced compared to NPU by 22.6\% along with 2$\times$ higher throughput. This trend suggests that GPU requires large matrices for complete resource utilization as the operation is compute-bound. 
 {For smaller matrices, the operation becomes memory-bound as the memory access overhead supersedes the computational requirement. NPU, due to the presence of DMA, achieves better performance than GPU. Therefore, NPU is preferred for smaller matrix multiplies for its better performance due to efficient memory access, and GPU is preferred for larger ones for better performance.}

\begin{figure}[!h]
    \centering
    \includegraphics[width=\linewidth]{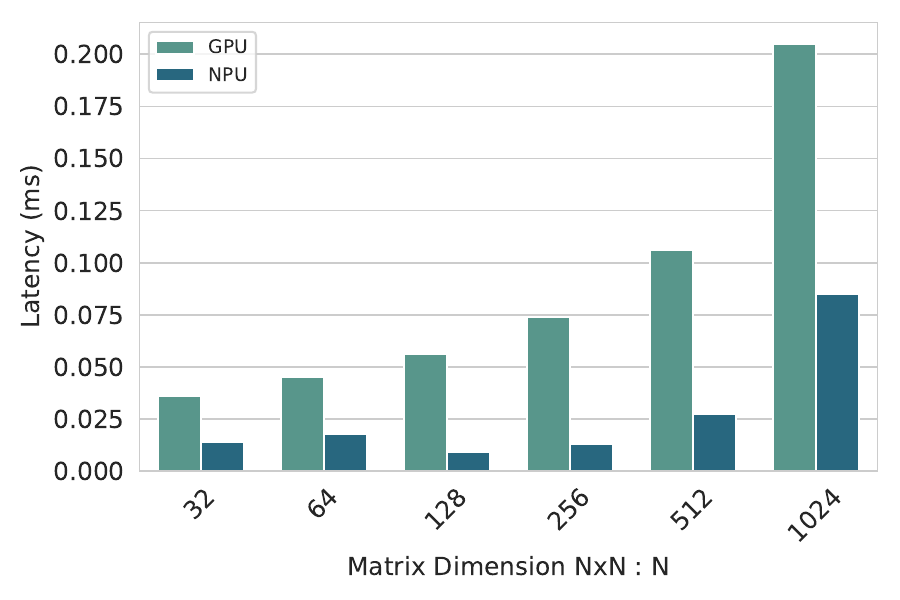}
    \caption{Latency for Matrix-Vector Multiplication}
    \label{latency_fig_la_mv}
\end{figure}

For matrix-vector multiplication, as shown in Fig. \ref{latency_fig_la_mv}, the NPU consistently outperforms the GPU, offering an approximate 2$\times$ speedup (58.54\% reduction in latency) across all matrix sizes. This is because the operation being memory bound {as data reuse is minimal, the DMA support in NPU} can efficiently control memory accesses in comparison to GPU. Interestingly, the NPU's latency values show a decrease when moving from matrix sizes of 64$\times$64 to 128$\times$128 before increasing again for larger sizes, which indicates that the NPU pipeline is optimized for matrices of size 128$\times$128. This performance characteristic indicates that NPU should be preferred over GPU for matrix-vector multiplication. Therefore, NPU should be the preference over GPU for matrix-vector multiplication operations.

\begin{figure}[!h]
    \centering
    \includegraphics[width=\linewidth]{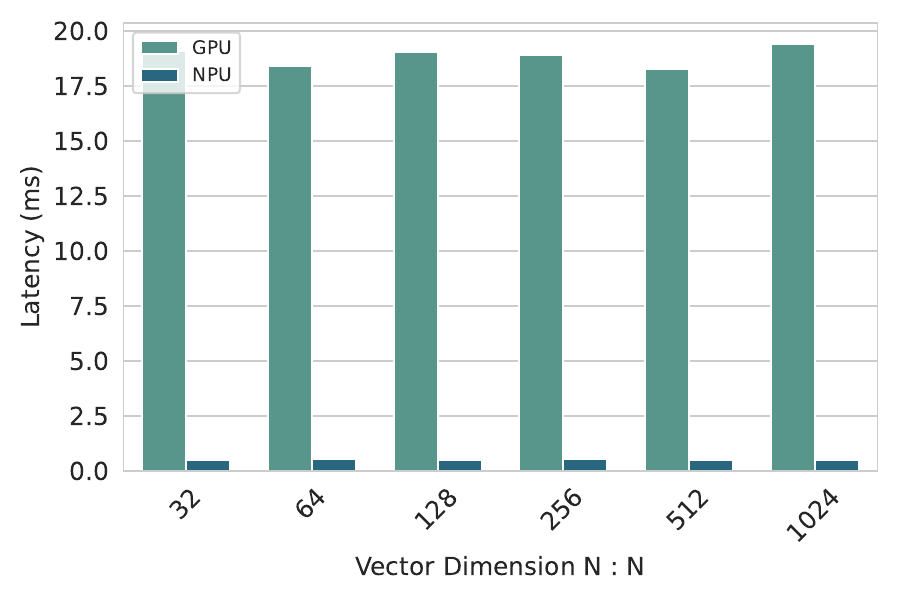}
    \caption{Latency for Dot Product}
    \label{latency_fig_dp}
\end{figure}

In the case of dot product operations, as shown in Fig. \ref{latency_fig_dp}, the NPU's performance is significantly better than that of the GPU. This substantial difference can be attributed to the {extremely limited data reuse in this operation, along with high synchronization overhead of the final sum, making it evidently a memory-bound problem that the NPU can efficiently handle in comparison to GPU. Another aspect of this operation is its simplicity and the highly sequential nature of computation, which makes it an ideal operation for the CPU, along with the fact that execution on the CPU eliminates additional memory accesses done by GPU and NPU. This explanation is well justified by the latency for dot-product as in Table ~\ref{tab: LA_performance}, which indicates CPU has lower latency than NPU.} Therefore, CPU is the choice over GPU and NPU for dot product operations.

\begin{figure}[!h]
    \centering
    \includegraphics[height=6cm, width=\linewidth]{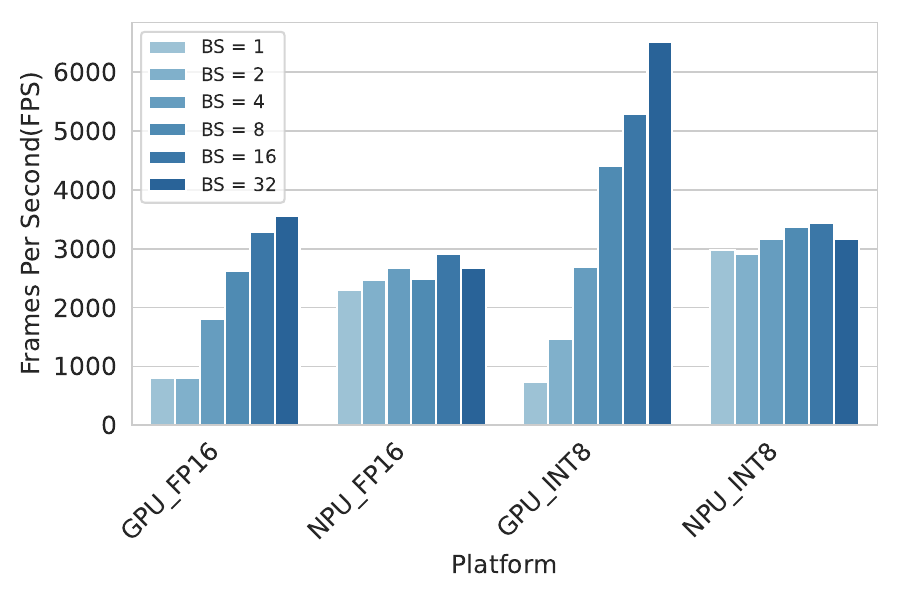}
    \caption{Throughput (FPS) with Varying Batch size for MobileNetV2 Video classification Inference}
    \label{throughput_fig_mn}
\end{figure}

Video classification using MobileNetV2 reveals interesting performance dynamics between NPU and GPU as shown in Table \ref{tab: Mobilenet_performance} and Fig. \ref{throughput_fig_mn}. For non-quantized FP-16 weights and a batch size of one, the NPU demonstrates nearly 3$\times$ lower inference latency and 3$\times$ higher throughput compared to the GPU. As batch size increases, the NPU's latency remains consistent while the GPU's latency decreases dramatically, eventually outperforming the NPU for larger batch sizes. {This result shows a trend that is very similar to that of matrix multiplication and emphasizes the analysis done previously. GPU performs the best in a compute-bound environment, which is established with higher batch sizes, resulting in increased resource utilization with a corresponding increase in throughput, and NPU performs the best in a memory-bound environment, which is the case with smaller batches.} When using quantized weights, the relative performance between NPU and GPU remains largely the same, with some nuances in performance for different batch sizes. These findings suggest that for consistent latency and classification with small batch sizes, inference on NPU is preferred, while GPU offers better performance with larger batch sizes. With the desire for consistent latency and for classification with small batch sizes, inference on NPU is preferred as compared to GPU, which only offers better performance with larger batch sizes.

\begin{table}
  \centering
  \caption{Inference Latency of Neural Network models}
  \begin{tabular}{|c|c|c|}
    \hline
    \textbf{Model} & \textbf{GPU} & \textbf{NPU}\\
    \hline
    MobileNetV2 & 4.95 ms & \textbf{1.73 ms}\\
    LSTM & \textbf{1.48 ms}& 4.10 ms\\
    TinyLlama & 8.30 sec & \textbf{2.49 sec} \\    
    \hline 
  \end{tabular}
  \label{tab: NN_performance}
\end{table}

For LSTM and LLM inference, we observe from Table \ref{tab: NN_performance} that LSTM, being a smaller model, exhibits low latency on both NPU and GPU. However, the GPU slightly outperforms NPU for LSTM inference. Conversely, for LLM inferencing, the NPU demonstrates remarkably efficient performance, nearly 4$\times$ better than the GPU.
{
This result aligns with our findings from matrix multiplication and matrix-vector multiplication tests. LLM inferencing primarily consists of prefill and decode stages, which largely rely on these operations respectively. Since the decode stage, dominated by matrix-vector multiplication, is executed multiple times compared to the single execution of the prefill stage, NPU emerges as the most suitable candidate for LLM inference. This is consistent with NPU's superior performance in matrix-vector operations.}
Based on these observations, NPU is the preferred choice for LLM inference in comparison to GPU. For LSTM inference, {GPU holds a slight edge in performance over NPU. Despite LSTM operations being primarily matrix-vector multiplies, NPU demonstrates degraded performance in comparison to GPU. 
This degradation can be attributed to the irregularity in memory accesses performed by LSTM.
The DMA in NPU can be effective in mitigating a memory-bound problem only when the accesses are regular; otherwise, it results in reduced efficiency as the DMA will have to undergo a new setup for every new access.} Therefore, for LLM inference, NPU is preferred, and for LSTM inference, GPU is preferred.

{
Our analysis reveals that NPU offers significant advantages in power efficiency compared to GPU for similar performance outcomes. While achieving comparable or better results across various tasks, the NPU consistently operates at less than half the peak power consumption of the GPU, drawing only 35 watts compared to the GPU's 75 watts. This substantial reduction in power usage makes NPU-based solutions particularly attractive for edge computing and mobile applications, where energy efficiency is crucial. The NPU's ability to deliver high performance at lower power consumption demonstrates its potential to enable more sustainable and cost-effective AI deployments in resource-constrained environments.
}

%% file: content/6_conclusion.tex
{
Our benchmark study on Intel AIPC, a heterogeneous platform with CPU, GPU, and NPU, revealed task-specific performance variations. NPU excelled in matrix-vector multiplication (58.54\% faster than GPU) and large language models (3.2$\times$ speedup). GPU outperformed in large matrix multiplications (22.6\% faster, 2$\times$ throughput) and LSTM networks (2.7$\times$ speedup). CPU was optimal for simple operations like dot product. For video classification, NPU performed better with small batch sizes, while GPU excelled with larger batches. We highlight the importance of strategic compute unit selection in heterogeneous computing for edge AI applications.

Our results can be extrapolated to other heterogeneous System-on-Chips with CPU/GPU/NPU compute units for several reasons. First, we utilize OpenVINO and PyTorch, which convert neural networks to an intermediate representation, ensuring portability across platforms. Second, OpenVINO's just-in-time compilation allows for performance consistency across hardware platforms with similar specifications. Third, the NPU's Streaming Hybrid Architecture Vector Engine architecture shares fundamental principles with other NPU designs, such as AMD's XDNA, Apple's Neural Engine, and Qualcomm's Hexagon, all optimized for parallel processing and efficient matrix operations. This architectural similarity suggests comparable performance characteristics across different NPU implementations in edge AI deployments.

Our results motivate the exploration of dual-compute solutions, where multiple {devices} can be leveraged simultaneously to maximize performance and energy efficiency. Future research directions include benchmarking energy consumption to optimize the performance-efficiency trade-off and extending our study to multi-platform heterogeneous computing. 
}